\documentclass[letterpaper, 10 pt, conference]{ieeeconf}  

\IEEEoverridecommandlockouts                              

\usepackage{graphicx}
\usepackage{capt-of}
\usepackage{balance}
\usepackage{booktabs}
\usepackage{amssymb}
\usepackage{bm}
\usepackage{amsmath,amssymb}
\usepackage{url}

\title{\LARGE \bf
SLIP-VLA: Single-Step Latent Imagination for Policy Learning in Vision-Language-Action Models
}

\author{
    Tianfu Li$^{1,*}$,
    Haoxuan Xu$^{2,*}$,
    Wenbo Chen$^{1,*}$,
    Haitian Li$^{3}$,
    Changchuan Yang$^{4}$,\\
    Xinhu Zheng$^{1}$,
    Jun Ma$^{1}$,
    Yuan Liu$^{2}$,
    Lujia Wang$^{1}$,
    Haoang Li$^{1}$%
    \thanks{$^{*}$Tianfu Li, Haoxuan Xu, and Wenbo Chen
    contributed equally to this work.}%
    \thanks{$^{1}$The Hong Kong University of Science and
    Technology (Guangzhou).}%
    \thanks{$^{2}$The Hong Kong University of Science and
    Technology.}%
    \thanks{$^{3}$Nanyang Technological University.}%
    \thanks{$^{4}$Zhejiang University.}%
}

\IEEEaftertitletext{%
    \begin{center}
        \includegraphics[width=\textwidth]{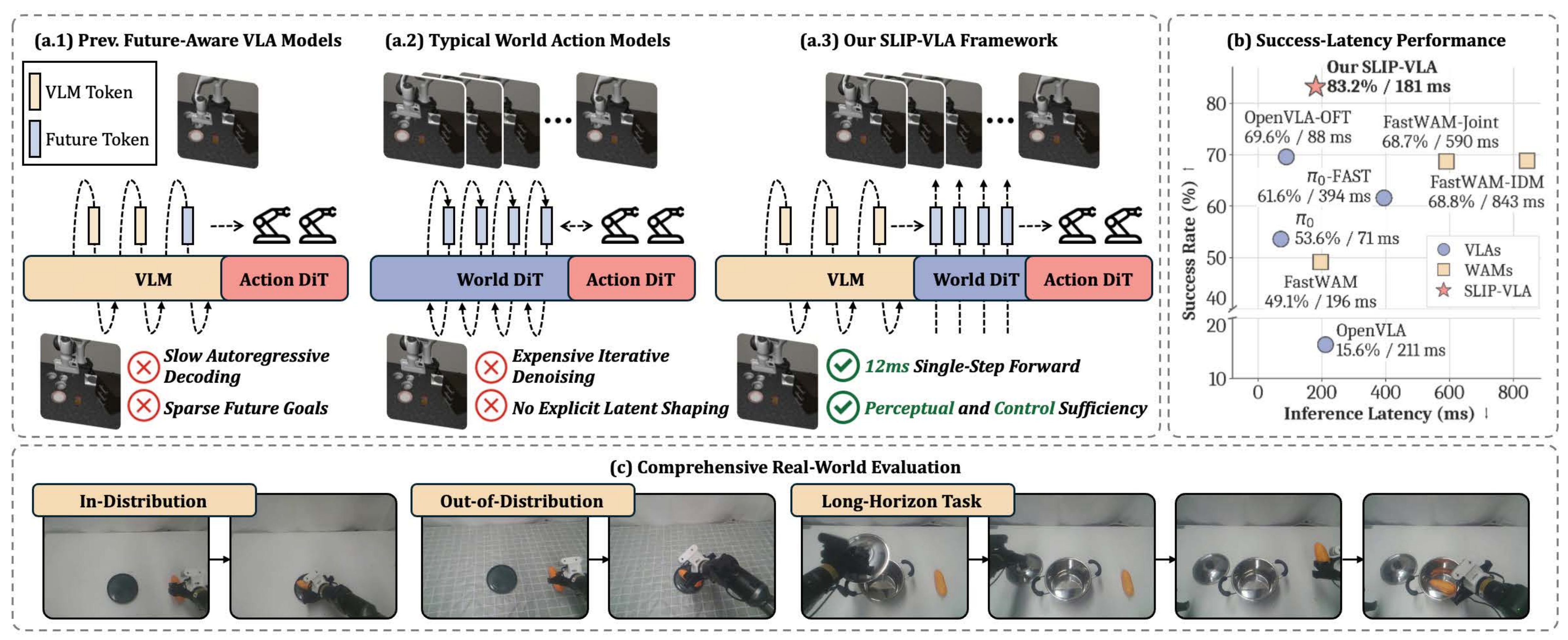}
        \captionof{figure}{
        \textbf{Overview of SLIP-VLA.}
        (a) Prior future-aware VLAs provide sparse future goals, while World Action Models capture temporally dense futures through costly iterative denoising.
        SLIP-VLA instead obtains temporally dense future latents in a single forward pass and explicitly shapes them for perceptual and control sufficiency.
        (b) On LIBERO-Plus, SLIP-VLA achieves the highest success rate among the plotted methods while maintaining competitive inference latency of 181 ms.
        (c) Real-world experiments demonstrate successful task execution across in-distribution, visual out-of-distribution, and long-horizon bimanual manipulation settings.
        }
        \label{fig:teaser}
    \end{center}
}

\begin{document}

\maketitle
\thispagestyle{empty}
\pagestyle{empty}

\begin{abstract}
Vision-Language-Action models are increasingly effective for robotic manipulation, yet most predict actions directly from current observations without explicitly modeling future scene evolution.
Recent methods introduce future prediction to improve action generation, but dense future modeling often requires expensive iterative denoising, while one-step alternatives can underperform their multi-step counterparts.
To reconcile efficient future modeling with strong action performance, we present \textbf{SLIP-VLA}, a policy learning framework that equips VLA models with a \textbf{Single-Step Latent Imagination} for future-aware action prediction. SLIP-VLA obtains temporally dense future latent representations with a single denoising update,
and we improve the \textbf{perceptual sufficiency} of these representations by aligning intermediate latents with future geometric and semantic features.
We further improve their \textbf{control sufficiency} through action-conditioned latent world modeling and inverse dynamics modeling, explicitly coupling latent transitions with robot actions.
Our SLIP-VLA achieves state-of-the-art performance across diverse simulation benchmarks and real-world manipulation tasks, while its single-step latent imagination takes only $12$ ms. Our project page and demonstration videos are available at \url{https://haoxuanxu1024.github.io/SLIP_VLA/}.
\end{abstract}

\section{INTRODUCTION}

Vision-Language-Action (VLA) models have made strong progress in robot manipulation, but most still predict actions directly from current observations~\cite{kim2024openvla,black2024pi0,black2025pi05}.
However, action generation often requires anticipating how objects, contacts, and task states will evolve beyond the current observation.
Recent VLA methods introduce predicted future images or motion as intermediate guidance, but their future reasoning is often based on slow autoregressive decoding and temporally sparse visual goals that capture selected future states rather than the full scene evolution~\cite{cen2025worldvla,zhong2025flowvla,zhao2025cotvla}, as illustrated in Fig.~\ref{fig:teaser}(a.1).
World Action Models (WAMs) shown in Fig.~\ref{fig:teaser}(a.2) extend this idea to temporally dense visual dynamics, but typically require expensive iterative denoising 
~\cite{motubrain2026,bi2026motus,li2026lingbotva}.
Although one-step variants reduce this cost, they can underperform their multi-step counterparts~\cite{ye2026dreamzero}.
This trade-off raises a key question: how should a single-step latent be shaped to expose sufficient future information for action prediction?

To support action generation, the future latent should encode sufficient perceptual structure, including the depth, spatial layout, and geometric boundaries of the future scene~\cite{wu2026geometryforcing,han2026gam,qian2026geopredict,wang2025vggt}.
Without these geometric cues, the policy may mislocalize contacts or target poses, leading to errors in grasping, alignment, and insertion.
Moverover, geometry helps localize potential interaction regions but does not identify which objects, parts, or states are relevant to the task.
Capturing this task-relevant semantics is challenging: generative objectives do not explicitly supervise it, and diffusion representations can lag behind strong self-supervised visual features~\cite{yu2025repa,leng2025repae,yang2023diffusionrep}.
Without such semantics, the policy may select the wrong target even when the future geometry is accurately represented~\cite{yu2026affordancevla,song2026reconvla}.
We refer to the joint requirements of these geometric and semantic cues as \textbf{perceptual sufficiency}.

Beyond preserving perceptual information, effective future latents should also capture how robot actions drive scene evolution.
A visually predictive latent may describe what the future scene looks like while discarding action-specific transition information, leaving the policy to infer action effects from data~\cite{huang2026a2world,tian2025pidm}.
Specifically, given the current latent and executed actions, the next latent state should be predictable, requiring the representation to preserve action-conditioned forward dynamics~\cite{zhou2025dinowm,maes2026lewm,huang2026a2world}.
Conversely, the transition between consecutive latent states should retain enough information to recover the executed actions, preventing the representation from ignoring the actions that produced the change~\cite{tian2025pidm,boylan2026contrastiveidm}.
We refer to these complementary forward and inverse requirements as \textbf{control sufficiency}.

To this end, we introduce \textbf{SLIP-VLA}: \textbf{S}ingle-Step \textbf{L}atent \textbf{I}magination for \textbf{P}olicy Learning in Vision-Language-Action Models.
SLIP-VLA first encodes the current visual observations and language instruction with a Vision-Language Model (VLM).
As shown in Fig.~\ref{fig:teaser}(a.3), a World DiT takes the VLM features and applies a single denoising update to produce temporally dense future latent representations, which condition the Action DiT.
For \textbf{perceptual sufficiency}, we align selected latent representations with future geometric and semantic features from pretrained visual encoders.
For \textbf{control sufficiency}, we regularize latent transitions with action-conditioned forward and inverse dynamics.
The auxiliary teachers and dynamics models are used only during training and are removed at inference.
Across diverse simulation benchmarks and real-world manipulation tasks, SLIP-VLA achieves state-of-the-art performance while its single-step latent imagination requires only $12$ ms at inference.
The resulting success--latency trade-off and evaluation on real-world tasks are demonstrated in Fig.~\ref{fig:teaser}(b, c).
Our contributions are summarized as follows:
\begin{itemize}

\item We introduce SLIP-VLA, a policy learning framework that equips VLA models with Single-Step Latent Imagination for future-aware action prediction, reconciling efficient future modeling with strong action performance;

\item We design perceptual sufficiency shaping by aligning intermediate representations with future geometric and semantic features, yielding future latents that better preserve task-relevant scene structure;

\item We design control sufficiency shaping through latent world modeling and inverse dynamics modeling, preserving action-dependent transitions and making the future latent more informative for robot control;

\item Our SLIP-VLA achieves SOTA performance across multiple simulation benchmarks and real-world tasks with the efficiency of single-step future imagination.

\end{itemize}





\section{RELATED WORK}

\noindent\textbf{Future-Aware VLAs.}
Recent VLA research has introduced future prediction as intermediate guidance for action generation~\cite{zhao2025cotvla}.
One line of work predicts future images as visual subgoals or indicators of subtask completion before decoding actions~\cite{zhao2025cotvla,long2026vista,wu2026streamvla,lu2026thinkingvla}.
Another represents anticipated scene changes through explicit motion or compact world features that encode dynamic, spatial, and semantic information~\cite{zhong2025flowvla,zhang2025dreamvla,wang2026lamp}.
Although these methods provide useful foresight, their predictions typically describe selected future states or compressed cues rather than the temporally dense scene evolution over the action horizon. These methods motivate temporally dense future representations, which are further explored by WAMs.

\noindent\textbf{Efficient Future Imagination in WAMs.}
WAMs jointly model robot actions and temporally dense visual dynamics, but generating the future trajectory typically requires expensive iterative denoising~\cite{motubrain2026,bi2026motus}.
To reduce this cost, some methods condition action generation directly on intermediate predictive representations rather than completing the generation process~\cite{li2026lingbotva,hu2024vpp,pai2025mimicvideo}.
Others further compress future imagination to a single denoising step or remove it entirely at inference~\cite{ye2026dreamzero,yuan2026fastwam}.
These studies show that useful control information can emerge before future generation is complete. However, one-step variants can still underperform their multi-step counterparts~\cite{ye2026dreamzero}.
This leaves open how an early predictive representation should be structured to remain useful for action prediction and control.

\noindent\textbf{Latent Representation Learning for Robot Control.}
The mentioned gap motivates a representation learning perspective: effective latent representations for robot control must support both scene understanding and action-conditioned prediction.
Representation alignment methods strengthen perceptual structure by aligning intermediate features with geometric or semantic representations from pretrained visual encoders~\cite{yu2025repa,leng2025repae,wu2026geometryforcing,li2026spatialforcing}.
However, perceptual richness alone does not ensure that a latent captures how robot actions induce future state changes.
Latent world models make this dependence explicit by predicting future representations from current states and robot actions~\cite{zhou2025dinowm,maes2026lewm,huang2026a2world}.
Inverse dynamics model provides a complementary constraint by recovering robot actions from latent state transitions, thereby encouraging the representation to preserve action-relevant information~\cite{tian2025pidm,boylan2026contrastiveidm}. These studies highlight the importance of preserving both task-relevant scene structure and action-dependent transitions in latent representations.
SLIP-VLA focuses on learning these properties under a single-step denoising update constraint, shaping the future representations for perceptual and control sufficiency.

\begin{figure*}[t]
    \centering
    \includegraphics[width=\textwidth]{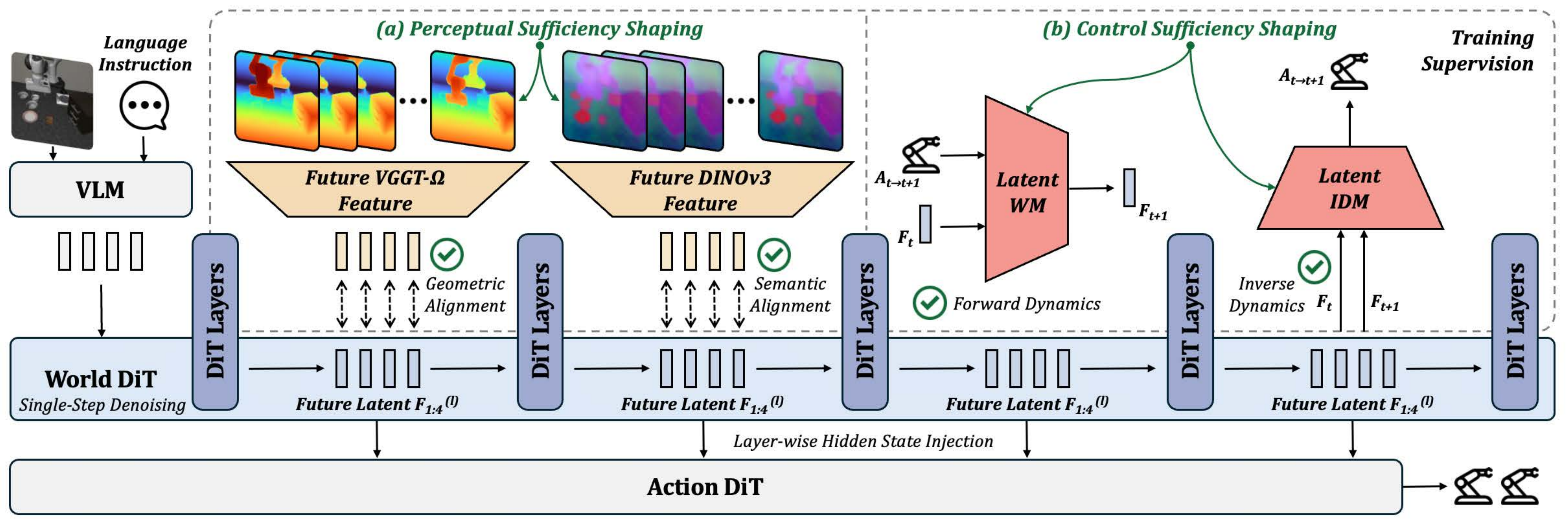}
    \caption{
    \textbf{Pipeline of SLIP-VLA.}
    The VLM backbone first encodes the multimodal context, and the World DiT then performs a single denoising pass to produce temporally dense future latents.
    (a) For perceptual sufficiency shaping, selected intermediate representations are aligned with future VGGT-$\Omega$ geometric features and DINOv3 semantic features.
    (b) For control sufficiency shaping, deeper latent representations are regularized through forward and inverse latent dynamics to preserve action-dependent transitions.
    The shaped future representations are injected layer-wise into the Action DiT for action generation, while all auxiliary shaping modules are used only during training.
    }
    \label{fig:method_overview}
    \vspace{-1.2em}
\end{figure*}

\section{METHODOLOGY} 

As illustrated in Fig.~\ref{fig:method_overview}, SLIP-VLA consists of a VLM, a World DiT, and an Action DiT.
The VLM encodes the current visual observations and language instruction.
Conditioned on the VLM features, the World DiT applies a single denoising update to produce future latent representations.
Selected World DiT representations are projected into the Action DiT feature space and used to condition the corresponding layers through layer-wise cross-attention. 

We first describe the construction of the single-step latent imagination, including the current-anchored source, the future latent slots, and the layer-wise conditioning of the Action DiT in Sec.~\ref{sec:single_step_latent_imagination}.
We then present perceptual sufficiency shaping through geometric and semantic alignment in Sec.~\ref{sec:perceptual_sufficiency_shaping}.
Next, we introduce control sufficiency shaping through latent forward and inverse dynamics in Sec.~\ref{sec:control_sufficiency_shaping}.
Finally, we describe the joint training objective and inference procedure in Sec.~\ref{sec:training_inference}.

\subsection{Single-Step Latent Imagination}
\label{sec:single_step_latent_imagination}

To preserve the current scene structure while leaving sufficient flexibility for future changes within a single denoising pass, our SLIP-VLA anchors the imagination source on the current observation rather than pure Gaussian noise.
We encode the current multi-view observation into a visual latent $\mathbf{z}_{\mathrm{cur}}$, repeat it across $K$ future slots, and construct the source latent as
\begin{equation}
    \mathbf{z}_{\mathrm{src}}
    =
    (1-\eta)\operatorname{Repeat}(\mathbf{z}_{\mathrm{cur}}, K)
    +
    \eta\widetilde{\epsilon},
\end{equation}
where $\widetilde{\epsilon}$ denotes Gaussian noise rescaled to the current latent scale and $\eta$ controls the noise ratio.
We use $K=4$ action-aligned future slots distributed along the prediction horizon.

With the current-anchored source $\mathbf{z}_{\mathrm{src}}$, the World DiT requires only a single denoising pass to obtain temporally dense predictive representations over all $K$ future slots.
Instead of using only the final World DiT output, we retain hidden representations from a set of selected layers $\mathcal{L}$.
For each layer $\ell\in\mathcal{L}$, we denote the resulting future latent sequence as
$\mathbf{F}_{1:K}^{(\ell)}=\{\mathbf{F}_{k}^{(\ell)}\}_{k=1}^{K}$,
where each $\mathbf{F}_{k}^{(\ell)}\in\mathbb{R}^{N\times D_w}$ preserves the spatial tokens of the $k$-th future slot.
These representations constitute our single-step latent imagination and are used directly for action generation without completing the denoising trajectory or decoding future RGB frames.
Finally, the Action DiT queries the projected future representations through layer-wise cross-attention. During training, we shape these same single-step future representations for perceptual and control sufficiency, so that the representations supplied to the Action DiT preserve task-relevant scene structure and action-dependent transitions.

\subsection{Perceptual Sufficiency Shaping}
\label{sec:perceptual_sufficiency_shaping}

To make the single-step latent perceptually sufficient, we align intermediate World DiT representations with complementary geometric and semantic features extracted from the future observations.
We apply the two objectives at different World DiT layers, using geometric supervision at an earlier layer $\ell_g$ and semantic supervision at a deeper layer $\ell_s$.

\subsubsection{Geometric Alignment}
We use a frozen VGGT-$\Omega$ encoder~\cite{wang2025vggt} to provide temporally contextualized geometric supervision.
For each camera view $v\in\mathcal{V}$, the current observation and all $K$ future frames are jointly encoded as a temporal window.
Let $\mathbf{o}_{0:K,v}$ denote this window, where $\mathbf{o}_{0,v}$ is the current frame and $\mathbf{o}_{1:K,v}$ are the future frames.
From the $m$-th selected VGGT-$\Omega$ layer, we construct the future geometric target as
\begin{equation}
    \mathbf{G}_{1:K,v}^{(m)}
    =
    \mathcal{R}_g
    \left(
        \left[
        T_g^{(m)}(\mathbf{o}_{0:K,v})
        \right]_{1:K}
    \right),
\end{equation}
where $T_g^{(m)}$ denotes the frozen teacher feature extractor and $\mathcal{R}_g$ resamples the teacher features to the World DiT spatial resolution.
Only the future-frame features are retained as supervision.

Let $\mathbf{F}_{1:K,v}^{(\ell_g)}$ denote the World DiT tokens associated with view $v$ at layer $\ell_g$.
A lightweight projector maps them to the corresponding teacher feature space:
\begin{equation}
    \widehat{\mathbf{G}}_{1:K,v}^{(m)}
    =
    P_g^{(m)}
    \left(
        \mathbf{F}_{1:K,v}^{(\ell_g)}
    \right).
\end{equation}

Following the geometric alignment objective, we preserve both feature direction and magnitude:
\begin{equation}
    \mathcal{L}_{\mathrm{geo}}
    =
    \frac{1}{M}
    \sum_{m=1}^{M}
    \left(
        \lambda_{\mathrm{ang}}
        \mathcal{L}_{\mathrm{ang}}^{(m)}
        +
        \lambda_{\mathrm{scale}}
        \mathcal{L}_{\mathrm{scale}}^{(m)}
    \right),
\end{equation}
with
\begin{equation}
    \mathcal{L}_{\mathrm{ang}}^{(m)}
    =
    -\mathbb{E}_{k,v,p}
    \left[
        \cos
        \left(
            \widehat{\mathbf{G}}_{k,v,p}^{(m)},
            \operatorname{sg}
            \left(
                \mathbf{G}_{k,v,p}^{(m)}
            \right)
        \right)
    \right],
\end{equation}
where $p$ indexes spatial tokens and $\operatorname{sg}(\cdot)$ denotes stop-gradient.
The scale term regresses the unnormalized geometric target from the normalized projected feature, preserving geometric magnitude information that is discarded by cosine alignment.

\subsubsection{Semantic Alignment}
Geometric supervision captures where future interactions occur, but does not explicitly preserve task-relevant object, part, and state semantics.
We therefore align a deeper World DiT representation with dense features from a frozen DINOv3 encoder~\cite{simeoni2025dinov3}.
Unlike VGGT-$\Omega$, DINOv3 processes each future frame independently.
For each future slot $k$ and view $v$, the semantic target is
\begin{equation}
    \mathbf{S}_{k,v}
    =
    \mathcal{R}_s
    \left(
        T_s(\mathbf{o}_{k,v})
    \right),
    \qquad k=1,\ldots,K,
\end{equation}
where $T_s$ is the frozen DINOv3 encoder and $\mathcal{R}_s$ resamples its dense patch features to the World DiT spatial grid.
A semantic projector maps the corresponding World DiT representation to the teacher feature space:
\begin{equation}
    \widehat{\mathbf{S}}_{k,v}
    =
    P_s
    \left(
        \mathbf{F}_{k,v}^{(\ell_s)}
    \right).
\end{equation}

We then optimize the dense feature alignment objective
\begin{equation}
    \mathcal{L}_{\mathrm{sem}}
    =
    -\mathbb{E}_{k,v,p}
    \left[
        \cos
        \left(
            \widehat{\mathbf{S}}_{k,v,p},
            \operatorname{sg}
            \left(
                \mathbf{S}_{k,v,p}
            \right)
        \right)
    \right].
\end{equation}

The resulting perceptual objective is
\begin{equation}
    \mathcal{L}_{\mathrm{perc}}
    =
    \lambda_{\mathrm{geo}}
    \mathcal{L}_{\mathrm{geo}}
    +
    \lambda_{\mathrm{sem}}
    \mathcal{L}_{\mathrm{sem}}.
\end{equation}

Together, these two alignment objectives encourage the single-step latent to preserve both temporally consistent geometric structure and rich semantic information.

\subsection{Control Sufficiency Shaping}
\label{sec:control_sufficiency_shaping}

Perceptually sufficient representations describe the predicted future, but do not necessarily capture how robot actions drive its evolution.
We therefore impose complementary forward and inverse dynamics objectives on deeper World DiT layers to explicitly couple future latent transitions with robot actions.
For two adjacent future slots $k$ and $k+1$, we denote by $\mathbf{A}_k$ the action segment executed between them.

\subsubsection{Forward Dynamics}
We first require the future latent evolution to be predictable from the executed robot actions.
At a World DiT layer $\ell_f$, a lightweight latent world model $W_{\theta}$ takes the current future latent and the intervening action segment to predict the next latent through a residual update:
\begin{equation}
\begin{aligned}
    \widehat{\mathbf{F}}_{k+1}^{+}
    &=
    \mathbf{F}_{k}^{(\ell_f)}
    +
    W_{\theta}
    \left(
        \mathbf{F}_{k}^{(\ell_f)},
        \mathbf{A}_{k}
    \right), \\
    \widehat{\mathbf{F}}_{k+1}^{-}
    &=
    \mathbf{F}_{k}^{(\ell_f)}
    +
    W_{\theta}
    \left(
        \mathbf{F}_{k}^{(\ell_f)},
        \mathbf{A}_{k}^{-}
    \right),
\end{aligned}
\end{equation}
where $\mathbf{A}_{k}^{-}$ is a mismatched action segment.
The next-slot representation $\mathbf{F}_{k+1}^{(\ell_f)}$ is treated as a stop-gradient target.
Let $d_k^{+}$ and $d_k^{-}$ denote the token-wise normalized feature distances from $\widehat{\mathbf{F}}_{k+1}^{+}$ and $\widehat{\mathbf{F}}_{k+1}^{-}$ to this target, respectively.
We define the forward-dynamics objective as
\begin{equation}
    \mathcal{L}_{\mathrm{fwd}}
    =
    \mathbb{E}_{k}
    \left[
        d_k^{+}
        +
        \lambda_{\mathrm{rank}}
        \left[
            m_{\mathrm{fwd}}
            + d_k^{+}
            - d_k^{-}
        \right]_{+}
    \right],
\end{equation}
where $[x]_{+}=\max(0,x)$.
The ranking term requires the correct action segment to explain the observed latent transition better than a mismatched one, discouraging the forward model from ignoring action information.

\subsubsection{Inverse Dynamics}
Forward dynamics alone does not guarantee that a latent transition preserves enough information to recover the action that caused it.
We therefore impose the complementary inverse relation at a deeper World DiT layer $\ell_i$.
A lightweight inverse dynamics model $I_{\psi}$ takes two consecutive dense latent states and predicts the intervening action segment:
\begin{equation}
    \widehat{\mathbf{A}}_{k}^{+}
    =
    I_{\psi}
    \left(
        \mathbf{F}_{k}^{(\ell_i)},
        \mathbf{F}_{k+1}^{(\ell_i)}
    \right).
\end{equation}
The model first aligns the two dense latent states through cross-attention and then uses learned action queries to decode the action sequence.
The action prediction error combines Smooth-L1 regression for robot motion and binary cross-entropy for the gripper state.

To ensure that the inverse dynamics model relies on the latent transition rather than the source state alone, we construct two negative pairs:
\begin{equation}
\begin{aligned}
    \widehat{\mathbf{A}}_{k}^{-}
    &=
    I_{\psi}
    \left(
        \mathbf{F}_{k}^{(\ell_i)},
        \mathrm{sg}
        \left(
            \mathbf{F}_{j}^{(\ell_i)}
        \right)
    \right), \\
    \widehat{\mathbf{A}}_{k}^{\mathrm{same}}
    &=
    I_{\psi}
    \left(
        \mathbf{F}_{k}^{(\ell_i)},
        \mathrm{sg}
        \left(
            \mathbf{F}_{k}^{(\ell_i)}
        \right)
    \right),
\end{aligned}
\end{equation}
where $\mathbf{F}_{j}^{(\ell_i)}$ is a mismatched future latent and $\mathrm{sg}(\cdot)$ denotes stop-gradient.
Let $e_k^{+}$, $e_k^{-}$, and $e_k^{\mathrm{same}}$ denote the action prediction errors of the positive, mismatched, and same-state pairs, respectively.
The inverse-dynamics objective is
\begin{equation}
\begin{aligned}
    \mathcal{L}_{\mathrm{inv}}
    =
    \mathbb{E}_{k}\big[
        &e_k^{+}
        +
        \lambda_{\mathrm{pair}}
        [m_{\mathrm{pair}} + e_k^{+} - e_k^{-}]_{+} \\
        &+
        \lambda_{\mathrm{same}}
        [m_{\mathrm{same}} + e_k^{+} - e_k^{\mathrm{same}}]_{+}
    \big].
\end{aligned}
\end{equation}

The mismatched pair requires the true future transition to be more informative than an unrelated one, while the same-state pair explicitly distinguishes a real state change from no transition.
Finally, we combine the two objectives as
\begin{equation}
    \mathcal{L}_{\mathrm{ctrl}}
    =
    \lambda_{\mathrm{fwd}}\mathcal{L}_{\mathrm{fwd}}
    +
    \lambda_{\mathrm{inv}}\mathcal{L}_{\mathrm{inv}}.
\end{equation}

Forward dynamics makes future evolution predictable from robot actions, whereas inverse dynamics makes the executed actions recoverable from latent evolution.
Together, they improve the control sufficiency of the single-step latent.

\subsection{Training and Inference}
\label{sec:training_inference}
During training, we jointly optimize future prediction, action generation, and the perceptual and control shaping objectives.
The World DiT uses a random-timestep branch to learn the full denoising trajectory and a shared first-step branch whose representations condition the Action DiT and receive perceptual and control shaping, without additional predictive forwards.
The Action DiT is trained with the standard conditional flow matching objective~\cite{black2024pi0}, conditioned on the VLM context and the selected single-step future representations.
The overall training objective is
\begin{equation}
\begin{aligned}
    \mathcal{L}
    =
    &\lambda_{\mathrm{act}}\mathcal{L}_{\mathrm{act}}
    +
    \lambda_{\mathrm{rand}}\mathcal{L}_{\mathrm{vid}}^{\mathrm{rand}}
    +
    \lambda_{\mathrm{first}}\mathcal{L}_{\mathrm{vid}}^{\mathrm{first}} \\
    &+
    \lambda_{\mathrm{perc}}\mathcal{L}_{\mathrm{perc}}
    +
    \lambda_{\mathrm{ctrl}}\mathcal{L}_{\mathrm{ctrl}},
\end{aligned}
\end{equation}
where $\mathcal{L}_{\mathrm{vid}}^{\mathrm{rand}}$ and
$\mathcal{L}_{\mathrm{vid}}^{\mathrm{first}}$ denote the video prediction losses from the random-timestep and first-step branches, respectively.
During early training, the timestep of the first-step branch is sampled within a small neighborhood of the source endpoint and gradually annealed to zero, progressively matching the inference condition.
We further adopt a staged optimization schedule that first stabilizes the predictive representation before jointly optimizing the full policy.

At inference, given the current observations, language instruction, and robot state, SLIP-VLA constructs the current-anchored source and performs a single World DiT forward to obtain multi-layer future representations.
The Action DiT directly queries these representations to generate the action chunk.
VGGT-$\Omega$, DINOv3, and the forward and inverse dynamics modules are training-only and removed at inference, introducing no additional online modules.

\section{EXPERIMENTS}

\begin{table}[!t]
    \caption{LIBERO success rates (\%). Best and second-best results are \textbf{bolded} and \underline{underlined}, respectively.}
    \vspace{-0.5em}
    \label{tab:libero_comparison}
    \centering
    \footnotesize
    \renewcommand{\arraystretch}{1.0}
    \begin{tabular*}{\columnwidth}{@{\extracolsep{\fill}}lccccc@{}}
        \toprule
        Method & Spatial & Object & Goal & Long & Avg. \\
        \midrule
        \textbf{\textit{General VLAs}} \\
        OpenVLA~\cite{kim2024openvla} & 84.7 & 88.4 & 79.2 & 53.7 & 76.5 \\
        OpenVLA-OFT~\cite{kim2025fine} & 97.6 & 98.4 & 97.9 & 94.5 & 97.1 \\
        $\pi_0$~\cite{black2024pi0} & 98.0 & 96.8 & 94.4 & 88.4 & 94.4 \\
        $\pi_{0.5}$~\cite{black2025pi05} & 98.8 & 98.2 & \underline{98.0} & 92.4 & 96.9 \\
        \midrule
        \textbf{\textit{Future-Aware VLAs}} \\
        CoT-VLA~\cite{zhao2025cotvla} & 87.5 & 91.6 & 87.6 & 69.0 & 83.9 \\
        FlowVLA~\cite{zhong2025flowvla} & 93.2 & 95.0 & 91.6 & 72.6 & 88.1 \\
        WorldVLA~\cite{cen2025worldvla} & 87.6 & 96.2 & 83.4 & 60.0 & 81.8 \\
        LaMP~\cite{wang2026lamp} & \underline{99.4} & \underline{99.8} & 97.4 & 96.7 & 98.3 \\
        \midrule
        \textbf{\textit{World Action Models}} \\
        Motus~\cite{bi2026motus} & 96.8 & \underline{99.8} & 96.6 & \underline{97.6} & 97.7 \\
        Fast-WAM~\cite{yuan2026fastwam} & 98.2 & \textbf{100.0} & 97.0 & 95.2 & 97.6 \\
        Fast-WAM-Joint~\cite{yuan2026fastwam} & \textbf{99.6} & 99.4 & \textbf{98.2} & 96.8 & \underline{98.5} \\
        Fast-WAM-IDM~\cite{yuan2026fastwam} & 98.8 & 97.8 & 97.8 & \underline{97.6} & 98.0 \\
        \midrule
        \textbf{SLIP-VLA (Ours)} & \textbf{99.6} & \underline{99.8} & \textbf{98.2} & \textbf{98.0} & \textbf{98.9} \\
        \bottomrule
    \end{tabular*}
    \vspace{-1.2em}
\end{table}

We conduct extensive experiments in simulation and the real world to answer the following questions:

\begin{itemize}
    \item \textbf{\textit{Q1}:} How does SLIP-VLA compare with state-of-the-art VLA and WAM methods on standard manipulation benchmarks?
    \item \textbf{\textit{Q2}:} Does the single-step latent imagination remain effective under distribution shifts and more challenging manipulation settings?
    \item \textbf{\textit{Q3}:} How do perceptual and control sufficiency shaping contribute to action performance and the learned future representations?
    \item \textbf{\textit{Q4}:} How much future denoising is necessary for action prediction, and what efficiency does single-step latent imagination provide?
\end{itemize}

\begin{table}[!t]
    \caption{LIBERO-Plus success rates (\%). Best and second-best results are \textbf{bolded} and \underline{underlined}, respectively.}
    \vspace{-0.5em}
    \label{tab:libero_plus_comparison}
    \centering
    \footnotesize
    \setlength{\tabcolsep}{0.5pt}
    \renewcommand{\arraystretch}{1.0}
    \begin{tabular*}{\columnwidth}{@{\extracolsep{\fill}}lcccccccc@{}}
        \toprule
        Method & Cam. & Robot & Lang. & Light & BG & Noise & Layout & Avg. \\
        \midrule
        \textbf{\textit{General VLAs}} \\
        OpenVLA~\cite{kim2024openvla} & 0.8 & 3.5 & 23.0 & 8.1 & 34.8 & 15.2 & 28.5 & 15.6 \\
        OpenVLA-OFT~\cite{kim2025fine} & 56.4 & 31.9 & 79.5 & 88.7 & 93.3 & 75.8 & 74.2 & 69.6 \\
        $\pi_0$~\cite{black2024pi0} & 13.8 & 6.0 & 58.8 & 85.0 & 81.4 & \underline{79.0} & 68.9 & 53.6 \\
        $\pi_0$-Fast~\cite{pertsch2025fast} & \underline{65.1} & 21.6 & 61.0 & 73.2 & 73.2 & 74.4 & 68.8 & 61.6 \\
        \midrule
        \textbf{\textit{Future-Aware VLAs}} \\
        WorldVLA~\cite{cen2025worldvla} & 0.1 & 27.9 & 41.6 & 43.7 & 17.1 & 10.9 & 38.0 & 25.0 \\
        LaMP~\cite{wang2026lamp} & 64.5 & \underline{69.6} & 88.2 & \textbf{95.3} & \textbf{97.4} & 76.9 & 73.8 & \underline{79.3} \\
        \midrule
        \textbf{\textit{World Action Models}} \\
        Fast-WAM~\cite{yuan2026fastwam} & 18.8 & 45.7 & 70.1 & 83.2 & 45.7 & 29.8 & 62.7 & 49.1 \\
        Fast-WAM-Joint~\cite{yuan2026fastwam} & 39.9 & 65.1 & \textbf{94.7} & 92.1 & 58.1 & 56.2 & 79.3 & 68.7 \\
        Fast-WAM-IDM~\cite{yuan2026fastwam} & 38.6 & 66.3 & \underline{94.2} & 87.7 & 58.0 & 58.8 & \textbf{81.5} & 68.8 \\
        \midrule
        \textbf{SLIP-VLA (Ours)} & \textbf{70.6} & \textbf{69.7} & 88.3 & \underline{94.1} & \underline{95.7} & \textbf{89.4} & \underline{81.4} & \textbf{83.2} \\
        \bottomrule
    \end{tabular*}
    \vspace{-0.3em}
\end{table}

\begin{table}[!t]
    \caption{RoboTwin~2.0 success rates (\%). Best and second-best results are \textbf{bolded} and \underline{underlined}, respectively.}
    \vspace{-0.5em}
    \label{tab:robotwin_comparison}
    \centering
    \footnotesize
    \renewcommand{\arraystretch}{1.0}
    \begin{tabular*}{\columnwidth}{@{\extracolsep{\fill}}lccc@{}}
        \toprule
        Method & Clean & Rand. & Avg. \\
        \midrule
        \textbf{\textit{General VLAs}} \\
        $\pi_0$~\cite{black2024pi0}
        & 65.9 & 58.4 & 62.2 \\
        $\pi_{0.5}$~\cite{black2025pi05}
        & 82.7 & 76.8 & 79.8 \\
        \midrule
        \textbf{\textit{Future-Aware VLAs}} \\
        WorldVLA~\cite{cen2025worldvla}
        & 42.5 & 32.2 & 37.4 \\
        \midrule
        \textbf{\textit{World Action Models}} \\
        Motus~\cite{bi2026motus}
        & 88.7 & 87.0 & 87.8 \\
        Fast-WAM~\cite{yuan2026fastwam}
        & \underline{91.9} & \textbf{91.8} & \textbf{91.8} \\
        Fast-WAM-Joint~\cite{yuan2026fastwam}
        & 90.8 & 90.3 & 90.6 \\
        Fast-WAM-IDM~\cite{yuan2026fastwam}
        & 91.2 & \underline{91.3} & \underline{91.3} \\
        \midrule
        \textbf{SLIP-VLA (Ours)}
        & \textbf{92.2} & \underline{91.3} & \textbf{91.8} \\
        \bottomrule
    \end{tabular*}
    \vspace{-1.2em}
\end{table}

\subsection{Experimental Setup}

\noindent\textbf{Benchmarks.}
We evaluate our SLIP-VLA on three simulation benchmarks and the real-world manipulation tasks.
LIBERO~\cite{liu2023libero} serves as the standard benchmark for multi-task manipulation, while LIBERO-Plus~\cite{fei2025liberoplus} evaluates zero-shot robustness under distribution shifts.
We further use RoboTwin~2.0~\cite{chen2026robotwin} to evaluate more complex bimanual manipulation across 50 tasks under the official settings.

\noindent\textbf{Implementation Details.}
All auxiliary supervision is derived from the same robot demonstrations used for policy learning, without additional demonstrations or manual annotations.
SLIP-VLA uses a Qwen3.5-2B VLM, a 12-layer World DiT, and a 24-layer Action DiT with primary and wrist camera inputs.
We use $K=4$ action-aligned future slots, $\eta=0.2$ for current-anchored initialization, and 16-step action chunks.
All models are trained on eight NVIDIA A100 GPUs.

\begin{figure}[!t]
    \centering
    \includegraphics[width=\columnwidth]{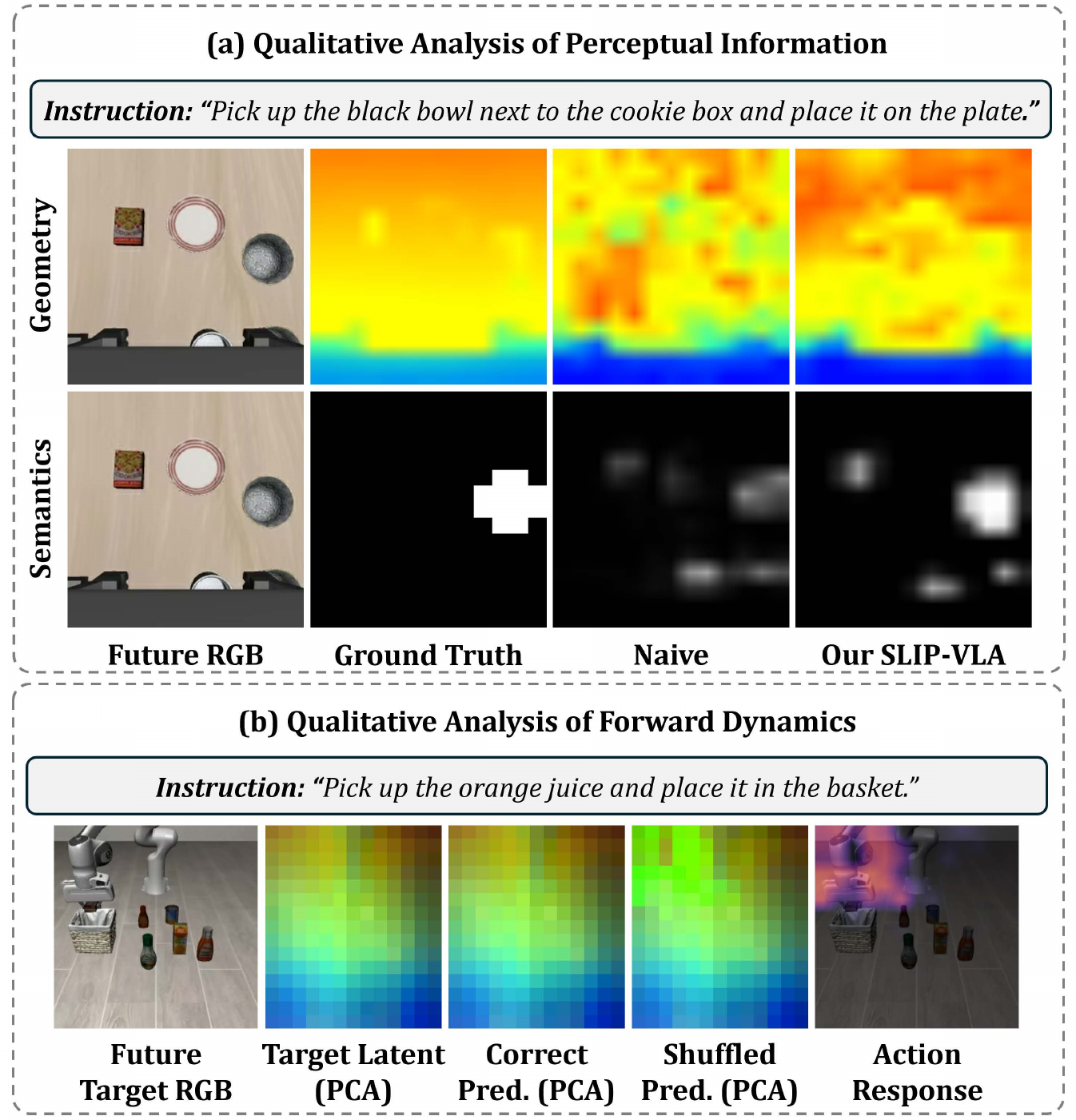}
    \caption{
    \textbf{Qualitative analysis} of perceptual and control information in the single-step latent.
    (a) Compared with the naive single-step baseline, SLIP-VLA recovers more coherent future geometry and more concentrated task-relevant semantic responses.
    (b) For forward dynamics, the correct action produces a prediction closer to the target latent, while a shuffled action induces clear spatial deviations; the action-response map highlights regions most affected by the action change.
    }
    \label{fig:latent_analysis}
\end{figure}

\subsection{Comparison with State-of-the-Art Methods (Q1 \& Q2)}

\noindent\textbf{LIBERO.}
Table~\ref{tab:libero_comparison} shows that SLIP-VLA achieves the highest average success rate, demonstrating that a shaped single-step latent can match or surpass strong future-aware and iterative WAM baselines on standard manipulation tasks.

\noindent\textbf{LIBERO-Plus.}
As shown in Table~\ref{tab:libero_plus_comparison}, SLIP-VLA achieves $83.2\%$ average success under distribution shift, outperforming the strongest compared method by $3.9$ points and the best WAM by $14.4$ points.
The larger margin highlights the benefit of latent shaping under out-of-distribution conditions.

\noindent\textbf{RoboTwin 2.0.}
As shown in Table~\ref{tab:robotwin_comparison}, SLIP-VLA achieves $91.8\%$ average success on the more complex bimanual tasks without external robot data pretraining, matching the best result.
Together with LIBERO-Plus, these results show that single-step latent imagination remains effective under both distribution shifts and more challenging manipulation.

\begin{figure}[!t]
    \centering
    \includegraphics[width=\columnwidth]{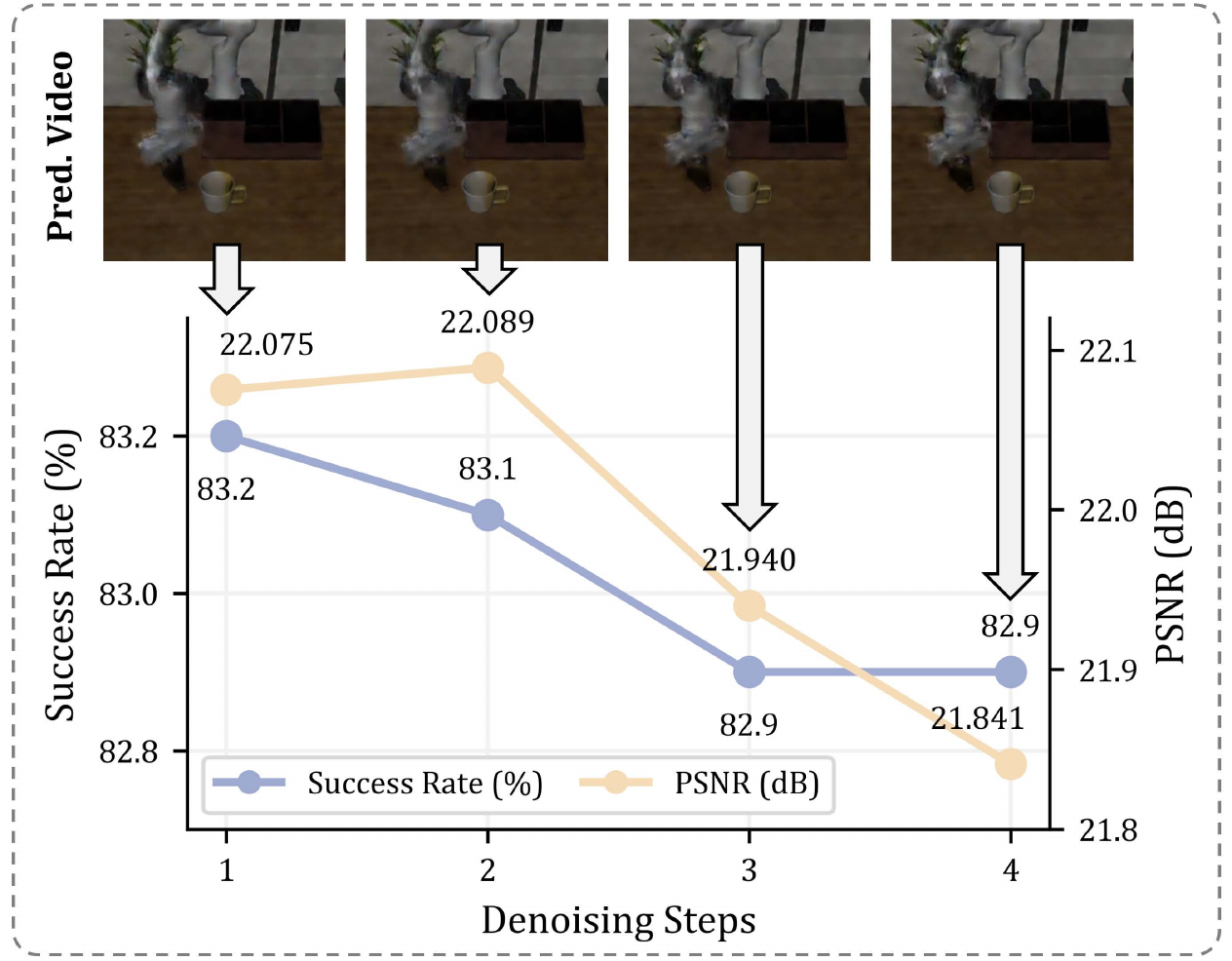}
    \caption{
    \textbf{Effect of denoising steps }on future prediction and control performance.
    The top row shows representative predicted videos with one to four denoising steps, while the curves report PSNR and LIBERO-Plus success rate.
    Additional denoising provides no consistent improvement in visual prediction quality and does not improve action performance, with the highest success rate achieved using a single step.
    }
    \label{fig:denoising_steps}
\end{figure}

\subsection{Ablation and Analysis (Q3 \& Q4)}

\begin{figure*}[!t]
    \centering
    \includegraphics[width=\textwidth]{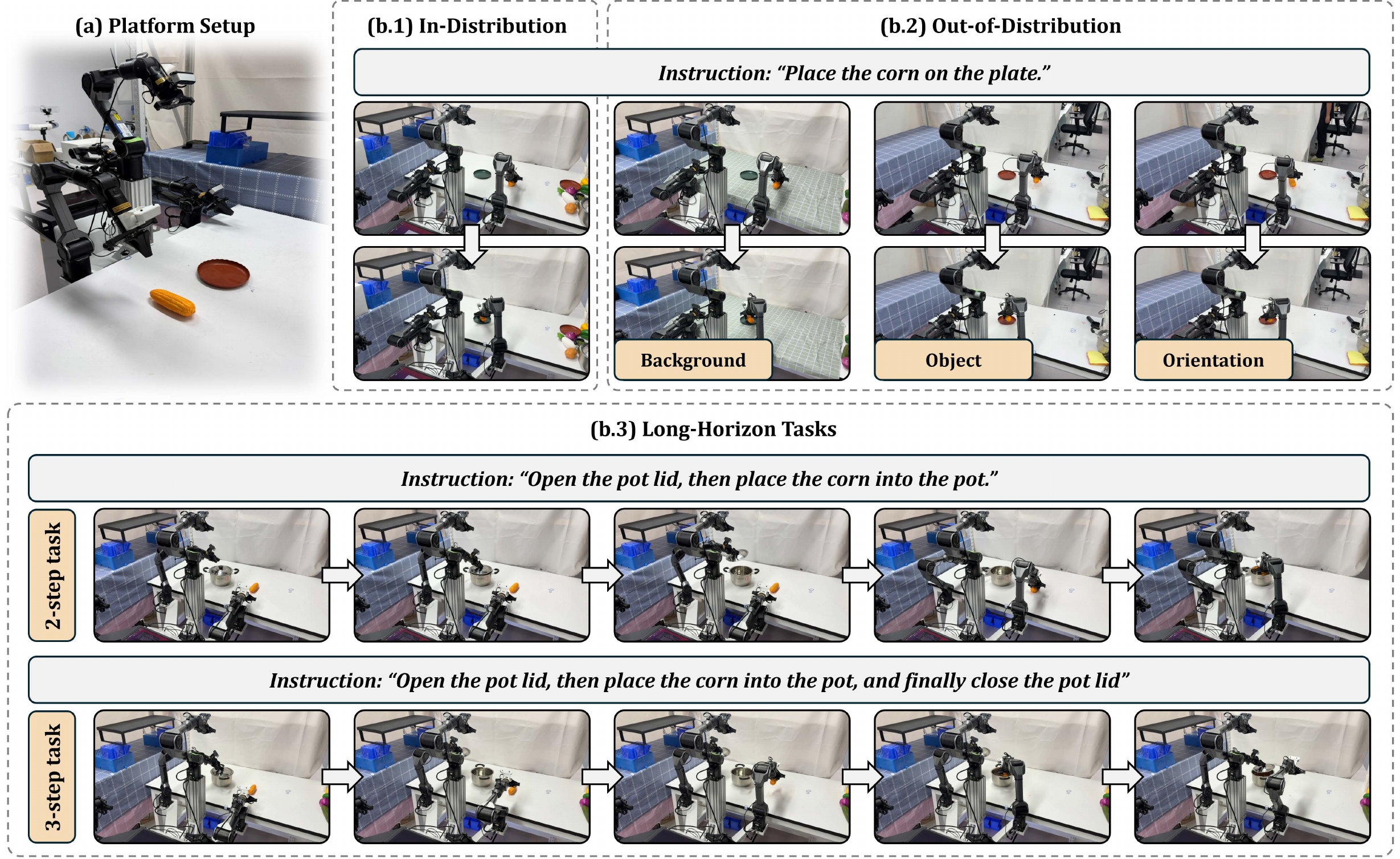}
    \caption{
    \textbf{Real-world evaluation of SLIP-VLA.}
    (a) The real-world manipulation platform.
    (b.1) Our SLIP-VLA successfully completes the in-distribution corn-to-plate task.
    (b.2) It remains successful under visual distribution shifts in background, object appearance, and object orientation.
    (b.3) SLIP-VLA further completes both two-step and three-step long-horizon tasks involving sequential interaction with the pot and the target object.
    }
    \vspace{-1.0em}
    \label{fig:real_world}
\end{figure*}


\noindent\textbf{Perceptual and Control Sufficiency Ablation.}
Table~\ref{tab:ps_cs_ablation} shows that both perceptual and control shaping improve over the naive single-step baseline, while their combination performs best on LIBERO and LIBERO-Plus.
The larger gain on LIBERO-Plus indicates that structured latent shaping is particularly beneficial under distribution shift.

\begin{table}[!t]
    \caption{Ablation study of perceptual and control sufficiency shaping.
    Success rates (\%) on LIBERO and LIBERO-Plus.}
    \vspace{-0.5em}
    \label{tab:ps_cs_ablation}
    \centering
    \footnotesize
    \renewcommand{\arraystretch}{1.05}
    \begin{tabular*}{\columnwidth}{@{\extracolsep{\fill}}lcc@{}}
        \toprule
        Variant & LIBERO & LIBERO-Plus \\
        \midrule

        Naive Single-Step
        & 97.7 & 79.0 \\

        \textit{w/} Perceptual Sufficiency
        & 98.3 & 81.4 \\

        \textit{w/} Control Sufficiency
        & 98.5 & 82.3 \\

        \midrule

        \textbf{\textit{w/} Perceptual + Control Sufficiency}
        & \textbf{98.9} & \textbf{83.2} \\

        \bottomrule
    \end{tabular*}
    \vspace{-1.2em}
\end{table}

\begin{table}[!t]
    \caption{Quantitative analysis of the perceptual sufficiency of frozen single-step latent representations. 
    }
    \vspace{-0.5em}
    \label{tab:perceptual_probe}
    \centering
    \footnotesize
    \renewcommand{\arraystretch}{1.05}
    \begin{tabular*}{\columnwidth}{@{\extracolsep{\fill}}lcccc@{}}
        \toprule
        & \multicolumn{2}{c}{Geometry} & \multicolumn{2}{c}{Semantics} \\
        \cmidrule(lr){2-3} \cmidrule(lr){4-5}
        Variant
        & Pearson $\uparrow$
        & AbsRel $\downarrow$
        & mIoU $\uparrow$
        & Dice $\uparrow$ \\
        \midrule

        Naive Single-Step
        & 0.804
        & 0.188
        & 0.388
        & 0.486 \\

        \textbf{SLIP-VLA (Ours)}
        & \textbf{0.916}
        & \textbf{0.112}
        & \textbf{0.590}
        & \textbf{0.683} \\

        \bottomrule
    \end{tabular*}
    \vspace{-0.3em}
\end{table}

\begin{table}[!t]
    \caption{\textbf{Quantitative analysis of control sufficiency in SLIP-VLA.}
    Forward dynamics measures latent prediction distance under different action perturbations, while inverse dynamics measures action recovery error under different latent-transition perturbations. Lower is better.}
    \vspace{-0.5em}
    \label{tab:control-analysis}
    \centering
    \footnotesize
    \renewcommand{\arraystretch}{1.05}
    \begin{tabular*}{\columnwidth}{@{\extracolsep{\fill}}lcc@{}}
        \toprule
        Evaluation Condition & Distance $\downarrow$ & $\Delta$ vs. Correct $\uparrow$ \\
        \midrule

        \textbf{\textit{Forward Dynamics}} \\
        Correct Action
        & \textbf{0.012} & -- \\

        Gripper Shuffled
        & 0.527 & 0.515 \\

        Motion Shuffled
        & 0.772 & 0.759 \\

        Fully Shuffled
        & 0.876 & {0.864} \\

        \midrule

        \textbf{\textit{Inverse Dynamics}} \\
        Correct Pair
        & \textbf{0.004} & -- \\

        Same-State
        & 1.747 & 1.743 \\

        Source-Only
        & 1.881 & 1.877 \\

        Next-Only
        & 1.962 & 1.958 \\

        Shuffled-Next
        & 2.124 & {2.120} \\

        \bottomrule
    \end{tabular*}
    \vspace{-1.2em}
\end{table}

\noindent\textbf{Perceptual Analysis of the Latent Representation.}
To evaluate perceptual sufficiency, we discard the training-time alignment heads and fit linear probes to frozen World DiT representations for future depth and task-relevant object masks.
Table~\ref{tab:perceptual_probe} shows improved geometry (higher Pearson and lower aligned AbsRel) and semantics (higher object mIoU and Dice), indicating that both are more accessible in the single-step latent.
The qualitative results in Fig.~\ref{fig:latent_analysis}(a) show the same trend.

\noindent\textbf{Control Analysis of the Latent Representation.}
To evaluate control sufficiency, we test whether latent transitions support action-conditioned future prediction and action recovery.
Table~\ref{tab:control-analysis} shows that shuffling actions worsens forward prediction, while corrupting latent transitions degrades inverse dynamics, revealing bidirectional action dependence.
Fig.~\ref{fig:latent_analysis}(b) further shows structured spatial deviations in the predicted future latent when the action changes.

\noindent\textbf{Source Initialization.}
Table~\ref{tab:source_initialization} shows that initializing future slots from the current latent outperforms pure Gaussian noise, highlighting the benefit of preserving current scene structure.
Adding a small amount of noise further improves performance by allowing future changes while retaining this structural prior.

\begin{table}[!t]
    \caption{Ablation study of source initialization strategies.
    Success rates (\%) on LIBERO and LIBERO-Plus.}
    \vspace{-0.5em}
    \label{tab:source_initialization}
    \centering
    \footnotesize
    \renewcommand{\arraystretch}{1.05}
    \begin{tabular*}{\columnwidth}{@{\extracolsep{\fill}}lcc@{}}
        \toprule
        Source Initialization & LIBERO & LIBERO-Plus \\
        \midrule

        Gaussian Noise
        & 98.6 & 82.5 \\

        Current Latent
        & 98.8 & 82.9 \\

        \textbf{Current + Noise (Ours)}
        & \textbf{98.9} & \textbf{83.2} \\

        \bottomrule
    \end{tabular*}
    \vspace{-1.5em}
\end{table}






\begin{table}[!t]
    \caption{
    Real-world success rates (\%).
    Each task condition is evaluated over 50 trials.
    }
    \vspace{-0.5em}
    \label{tab:real_world}
    \centering
    \footnotesize
    \renewcommand{\arraystretch}{1.05}
    \begin{tabular*}{\columnwidth}{@{\extracolsep{\fill}}lcc@{}}
        \toprule
        Task & $\pi_0$ & \textbf{SLIP-VLA (Ours)} \\
        \midrule

        \textbf{\textit{In-Distribution}} \\
        Corn $\rightarrow$ Plate
        & 72 & \textbf{86} \\

        \midrule
        \textbf{\textit{Out-of-Distribution}} \\
        Unseen Background
        & 58 & \textbf{70} \\ 
        Unseen Object  Appearance
        & 54 & \textbf{70} \\ 
        Unseen Object Orientation
        & 52 & \textbf{66} \\ 

        \midrule
        \textbf{\textit{Long-Horizon}} \\
        Open $\rightarrow$ Corn into Pot (2-step)
        & 68 & \textbf{82} \\
        Open $\rightarrow$ Corn into Pot $\rightarrow$ Close (3-step)
        & 60 & \textbf{74} \\ 

        \bottomrule
    \end{tabular*}
    \vspace{-1.2em}
\end{table}


\noindent\textbf{Denoising Steps and Efficiency.}
Fig.~\ref{fig:denoising_steps} shows that additional denoising brings no consistent gain in visual quality or LIBERO-Plus performance, with the highest success rate achieved in a single step.
This answers \textbf{\textit{Q4}}: once properly shaped, the single-step latent is sufficient for action prediction, eliminating unnecessary iterative refinement.


\subsection{Real-World Experiments}
\label{sec:real_world}

\noindent\textbf{Platform.}
We conduct real-world experiments on the AgileX COBOT Magic platform shown in Fig.~\ref{fig:real_world}(a).
COBOT Magic is a bimanual manipulation platform equipped with PiPER robotic arms, a TRACER mobile base, two wrist cameras, and a top camera.
Our policy uses the multi-view visual observations to perform closed-loop manipulation on the physical robot.

\noindent\textbf{Task Design.}
As illustrated in Fig.~\ref{fig:real_world}(b.1)--(b.3), we evaluate three groups of tasks with increasing difficulty.
The in-distribution (In-D) setting evaluates the standard corn-to-plate task.
For out-of-distribution (OOD) evaluation, we preserve the same manipulation objective while independently introducing unseen backgrounds, objects, and object orientations.
We further consider two long-horizon tasks: a two-step task that opens the pot lid and places the corn into the pot, and a three-step task that additionally requires closing the lid after placement.
Each condition is evaluated over 50 trials.

\noindent\textbf{Results.}
As shown in Table~\ref{tab:real_world}, SLIP-VLA consistently outperforms $\pi_0$ across all six real-world conditions, with absolute improvements of $12$--$16$ percentage points.
On the In-D task, SLIP-VLA improves the success rate from $72\%$ to $86\%$.
The advantage persists across unseen backgrounds, objects, and orientations, where SLIP-VLA achieves $70\%$, $70\%$, and $66\%$ success, respectively.
It also maintains strong performance as the task horizon increases, achieving $82\%$ on the two-step task and $74\%$ on the more challenging three-step task.
The successful rollouts in Fig.~\ref{fig:real_world}(b.1)--(b.3) further demonstrate that single-step latent imagination transfers effectively to physical manipulation under visual distribution shifts and extended action sequences.



\section{CONCLUSION}

We presented SLIP-VLA, a policy learning framework that replaces iterative future imagination with a single-step predictive latent for efficient future-aware action prediction.
SLIP-VLA shapes this latent for perceptual sufficiency through geometric and semantic alignment, and for control sufficiency through forward and inverse dynamics.
These training-time objectives introduce no additional modules at inference.
Across simulation benchmarks and real-world manipulation tasks, SLIP-VLA achieves state-of-the-art performance while requiring only $12$ ms for single-step latent imagination.
Our results show that effective future-aware control does not require fully denoised visual predictions when the predictive latent is properly shaped.



\balance
\bibliographystyle{IEEEtran}
\bibliography{references}
\end{document}